\documentclass{article}

\usepackage{PRIMEarxiv}
\usepackage[font=bf]{caption}
\usepackage[utf8]{inputenc} 
\usepackage[T1]{fontenc}    
\usepackage{hyperref}
\hypersetup{colorlinks=true,urlcolor=black, citecolor=black, linkcolor=black} 
\usepackage{url}            
\usepackage{booktabs}       
\usepackage{amsfonts}       
\usepackage{nicefrac}       
\usepackage{microtype}      
\usepackage{lipsum}
\usepackage{fancyhdr}       
\usepackage{graphicx}       
\usepackage{xcolor}
\usepackage{subcaption}
\usepackage[numbers]{natbib}
\usepackage{tikz}
\usetikzlibrary{calc,trees,positioning,arrows,chains,shapes.geometric,%
    decorations.pathreplacing,decorations.pathmorphing,shapes,%
    matrix,shapes.symbols, positioning, mindmap}   
\usepackage{rotating}
\usepackage{makecell}
\usepackage{multirow}
\usepackage{tabularx}
\newcolumntype{Y}{>{\hsize=.28\hsize}X}
\usepackage{booktabs}
\usepackage{arydshln}
\usepackage{utfsym}
\usepackage{comment}
\usepackage{microtype}
\usepackage{xcolor}
\usepackage{colortbl}
\definecolor{light-gray}{gray}{0.65}
\usepackage{fontawesome}

\newcommand{\user}{\faUser}
\newcommand{\both}{\faUserPlus}

\usepackage{amsmath}    

\usepackage{todonotes}[disable] 

\title{Which PPML Would a User Choose? A Structured Decision Support Framework for Developers to Rank PPML Techniques Based on User Acceptance Criteria
}

\author{
  \begin{minipage}[t]{0.45\textwidth}
    \centering
    \textbf{Sascha Löbner}, \textbf{Sebastian Pape} \\ 
    {\normalfont Goethe University \\ 
    Frankfurt am Main, 
    Germany \\} 
    \texttt{\{sascha.loebner, sebastian.pape\}@m-chair.de} 
  \end{minipage}%
  \hfill
  \begin{minipage}[t]{0.45\textwidth}
    \centering
    \textbf{Vanessa Bracamonte}, \textbf{Kittiphop Phalakarn} \\ 
    {\normalfont KDDI Research, Inc. \\ 
    Saitama, 
    Japan \\} 
    \texttt{\{va-bracamonte, xki-phalakarn\}@kddi-research.jp} 
  \end{minipage}
}

\begin{document}
\maketitle

\begin{abstract}
Using Privacy-Enhancing Technologies (PETs) for machine learning often influences the characteristics of a machine learning approach, e.g., the needed computational power, timing of the answers or how the data can be utilized. When designing a new service, the developer faces the problem that some decisions require a trade-off. For example, the use of a PET may cause a delay in the responses or adding noise to the data to improve the users' privacy might have a negative impact on the accuracy of the machine learning approach. As of now, there is no structured way how the users' perception of a machine learning based service can contribute to the selection of Privacy Preserving Machine Learning (PPML) methods. This is especially a challenge since one cannot assume that users have a deep technical understanding of these technologies. Therefore, they can only be asked about certain attributes that they can perceive when using the service and not directly which PPML they prefer.

This study introduces a decision support framework with the aim of supporting the selection of PPML technologies based on user preferences. Based on prior work analysing User Acceptance Criteria (UAC), we translate these criteria into differentiating characteristics for various PPML techniques. As a final result, we achieve a technology ranking based on the User Acceptance Criteria while providing technology insights for the developers. We demonstrate its application using the use case of classifying privacy-relevant information.

Our contribution consists of the decision support framework which consists of a process to connect PPML technologies with UAC, a process for evaluating the characteristics that separate PPML techniques, and a ranking method to evaluate the best PPML technique for the use case.

\end{abstract}


\section{Introduction}\label{sec:intro}
Which Privacy Preserving Machine Learning (PPML) technique a user would prefer to be implemented in a specific use case is a difficult question to assess from a developer's perspective. Since users lack fundamental knowledge about these technologies it is difficult to obtain preferences for implementing a PPML technique in a certain application.  But for the actual use of an application by the user, meeting their needs is a crucial requirement for AI service providers. This is on the one hand driven by the need to meet legal regulations such as the GDPR. On the other hand, privacy can be used as a selling point when comparing to other competitors. Although the principles of privacy by design~\citep{cavoukian2009privacy} claim full functionality in a win-win approach, in practice this is not easy to achieve for Machine Learning (ML) applications. Almost always PPML techniques such as Differential Privacy (DP), Homomorphic Encryption (HE) or Secure Multiparty Computation (SMPC), come with different trade-offs in performance, accuracy and adaptability~\citep{loebnerTB}. Moreover, these techniques are often not dominating each other but come with individual trade-offs that influence different aspects of an application. 
To address this research gap, we provide a structured decision support framework for developers to support them in the selection of a PPML technology that best suits the users' preferences. The result can be used on its own or in a more general PPML selection process as one decision criterion out of many. To achieve this, we built on the analysis of \citep{loebnerTB} who elicited User Acceptance Criteria (UAC) that influence the users' acceptance of an application based on frequently used models, such as IUIPC \citep{zeng2020all, malhotra2004internet} or APCO \citep{benamati2017empirical} model. We use our framework to translate these criteria into differentiating PPML Characteristics among PPML technologies.

\par
Our contribution is the provision of a structured decision support framework for developers to rank PPML techniques based on UAC to answer the research question (RQ): “Which PPML technique would a User Choose?''. To better address this RQ, we split it into three research objectives (RO). 
\begin{itemize}
    \item RO1: Providing a process on how to connect PPML technologies with UAC. We contribute by providing precise formulas to apply the suggested mapping of \citet{loebnerTB} to calculate a PPML Characteristic preference score from UAC. 
    \item RO2: Providing a process for evaluating the characteristics that separate PPML techniques. We describe how PPML Characteristics can be divided into categories. Moreover, we suggest setting weights for each category using weight vectors.
    \item RO3: Providing a ranking method to evaluate the best PPML technique for the use case, based on RO1 and RO2. We contribute by providing a formula that connects PPML Characteristic preference scores translated from the user input and the weighted PPML categories. 
\end{itemize}

We address our research objectives by guiding the reader through all the steps of our framework. Furthermore, we showcase the use of the framework by the example of a simplified Privacy Sensitive Information (PSI) detection application. Thus, the focus is on the process not on the evaluation of the mapping itself. To be precise, based on the showcasing, we provide an explanation for the PSI detection application of how to arrive with our framework at the weighted list of PPML techniques, utilising the users' UAC as an input to the framework.
\par
\newcommand{\sectname}[1]{(\textit{#1})}
The structure of this paper is as follows: In section \ref{sect:relwork} \sectname{related literature} we collect papers that provide decision support to choose PPML techniques. Section \ref{sec:method} \sectname{methodology}  describes relevant entities, research objectives, and the methodology used to address them. In section \ref{sec:framew_descr} \sectname{framework description} we explain the calculation steps and how to apply them, which is the major contribution of this paper. In section \ref{sec:scenario}, we introduce a \textit{use case: PSI detection in texts}, which we use as an example to demonstrate the use of the framework. In section \ref{sec:application} \sectname{framework application} we apply the steps from section \ref{sec:framew_descr} to the use case described in section \ref{sec:scenario}. 
In section \ref{sec:discussion} \sectname{discussion}  we reflect on our results and implementation considerations taking into account the limitations and suggesting avenues for future research. Finally, section \ref{sec:conclusion} \sectname{conclusion} summarises our main findings.

\section{Related Literature}\label{sect:relwork}
There are several studies that provide classification and comparison of PPML techniques. \citet{zheng2019challenges} classified PPML systems based on key technologies with a special focus on applying the PPML techniques to IoT. Thus, they compared the techniques according to computation and communication overhead. They first considered whether the system is for training or for inference. For privacy-preserving training, they further divided into parameter transmission-based techniques (e.g., federated learning) and data transmission-based techniques (e.g., anonymization and data obfuscation). From their classification, it is difficult to see which key technologies can be combined. \citet{tanuwidjaja2020privacy} presented a similar classification of privacy-preserving deep learning on machine learning as a service.
\citet{boulemtafes2020review} classified PPML systems based on the structures of the systems. They divided the system settings into four layers: (1) learning, inference, or releasing a model, (2) collaborative learning (multiple participants are involved) or individual learning (a single participant is involved), (3) server-based (most tasks are outsourced) or server-assisted (tasks are performed cooperatively between participants and the outsourcing servers), and (4) key technological concepts (e.g., encryption). Similar to \cite{zheng2019challenges}, it is difficult to see from their classification which key technologies can be combined. For comparison of PPML techniques, they considered effectiveness (e.g., accuracy), efficiency (e.g., running time), and privacy.

\citet{tran2021efficient} compare different approaches such as HE, DP, SMPC and FL for privacy preserving decentralized deep learning and evaluate different artifacts such as existence of bottlenecks, privacy, utility reduction, training latency or performance costs. 

\citet{xu2021privacy} provided a framework explaining at which level different PPML techniques can be combined. This work may be considered as a combination of \cite{zheng2019challenges}, \cite{tanuwidjaja2020privacy}, and \cite{boulemtafes2020review}. For the first category, PPML systems are divided by phases: data preparation, model training, and model serving (deployment and inference). For the second category, PPML systems are divided according to privacy guarantee: object-oriented (data and models) and pipeline-oriented (boundary and trust assumption). For the last category, PPML systems are divided according to technical utility: data publishing approaches (elimination-based, perturbation-based, and confusion-based), data processing approaches (additive mask, garbled circuits, modern cryptographic, mixed-protocol, and trusted execution environment), architecture approaches (delegation-based ML, distributed selective SGD, federated learning, and knowledge transfer), and hybrid approaches. 
\par
\citet{loebnerTB} propose a mapping to translate UAC into PPML Characteristics by eliciting influencing relationships through joint coding and expert interviews. However, they do not present how to calculate the technology that suits an application best.
\par
To the best of our knowledge there exists no research explaining how user acceptance criteria can be used to elicit which PPML technology fits best to an application.


\section{Methodology}\label{sec:method}

This section presents involved entities  and the methodology addressing our research objectives.

\subsection{Entities Involved}
We build on the entities that were already introduced by\;\citet{loebnerTB}. Therefore, we only provide a brief summary of our definitions.


\par
\textit{Data entity:} A natural person who is providing private data for training a PPML model. No computation results are provided to the \textit{data entity}.
\par
\textit{User:} A natural person using an AI service as a customer, providing private input data and receiving private results. \textit{User} and \textit{data entity} are data subjects as defined in the GDPR \citep{GDPR16}.
\par
\textit{Developer:} Commissioned by the AI service provider to create PPML models for users, using data from \textit{data entities} and \textit{user}. Developing a PPML model is not restricted to a single occurrence but can be done continuously. The required data to train a model is taken (a) from the data entity to e.g. built a pre-trained model or (b) from the user input.
\par
\textit{AI service provider:} Utilises the PPML model that is commissioned and build by the \textit{developer} in a user relationship. The \textit{AI service provider} takes over all communication with \textit{user} and \textit{data entity}.
\par
\textit{Experts:} Exhibit advanced background knowledge in PPML or user privacy and are important to keep the presented framework up to date.

\subsection{Research Objectives}
To support AI service providers and developers, we  aim to identify which PPML technique a user would prefer to be implemented for an ML service that requires the disclosure of personal data. Therefore, we focus in this line of research on providing a process to evaluate different PPML techniques based on the user preferences elicited by \citet{loebnerTB}. 
Because all PPML techniques can improve privacy but have different advantages and disadvantages a distinction is required by analysing technical differences. However, due to continuous technical improvement of PPML technologies we focus on providing a process on how to build an overall evaluation of eligible techniques (see section \ref{sec:framew_descr}). Thus our framework is robust against changes in technology in future. We showcase the framework's application in section \ref{sec:application} taking over the role of framework users, utilising existing research and prototypes on the PPML technologies eligible for the use case. Finally, our approach aims to provide a structured decision support for developers and service providers to identify the PPML techniques with good (expected) user acceptance. Thus, for this line in research we have split the ranking process of PPML techniques into three research objectives (RO):


\paragraph*{\textbf{RO1: Providing a process on how to connect PPML technologies with UAC}} 

We explain and showcase how UAC preference scores can be translated into PPML Characteristic scores. This is done by utilising the mapping provided in \citet{loebnerTB}. This mapping contains already expert validated connections between UAC and PPML Characteristics. While the survey can also be conducted by the AI service provider, the translation into the PPML Characteristic scores should happen within the AI developer to adjust the mapping to the use case, if necessary. 

\paragraph*{\textbf{RO2: Providing a process for evaluating the PPML characteristics that separate PPML techniques}}
We provide a process on how the PPML Characteristics that separate different PPML techniques can be evaluated, introducing PPML criteria and criteria weights. We also showcase how the PPML criteria can be set up. This fine-grained evaluation is expected to be performed within the AI service chain, e.g., as paid service of an experienced PPML developer or as paid consultancy. 

\paragraph*{\textbf{RO3: Providing a ranking method to evaluate the best PPML technique for the use case, based on RO1 and RO2}}
PPML Characteristic preference scores (RO1) elicited from the user input and the weighted PPML Characteristics (RO2)
are used to compute a PPML technology score that can be used for technology decision support. The unique contribution of this score is that it takes user preferences into account. This task can also be performed as a service within the AI service chain because background knowledge in PPML from e.g. similar projects is likely to speed up this process and increase the accuracy of the evaluation. 
\par

\subsection{Threat Model}\label{sec:threat_model}  
To evaluate the level of protection of the PPML techniques, it is crucial to define an individual threat model for the application in focus. 
Since the threat model relies heavily on the ML application, i.\,e.\ its data input and output and the aim of the application, we cannot provide a specific threat model for each possible application. To still allow our framework to consider attacks, we build a somewhat generic threat model by considering a set of common attacks that are structured based on \citet{li2020federated} in local and global privacy. A model protected against third parties except members of the AI service chain is considered globally private, (\textit{resilience against attacks}) and it is  locally private if protected against all entities, including internal adversaries (\textit{purpose and access limitation}). 
We assume that attackers who are not part of the AI service chain, do not possess any information about the data or the model and have no physical access to the infrastructure. They can only interact with the target model by providing inputs and receiving a predicted result. 
\par
We assume that attackers inside the AI service chain, are able to collect information about the data and model over time if no protection mechanism prevents this. They might possess additional information about the user or data entity. The attacker has physical access to the infrastructure. The attackers have full access to the target model and its prediction if not prevent by PPML.

\section{Framework Description}\label{sec:framew_descr}

\begin{figure}[htb!]
\centering
\includegraphics[width=1\textwidth]{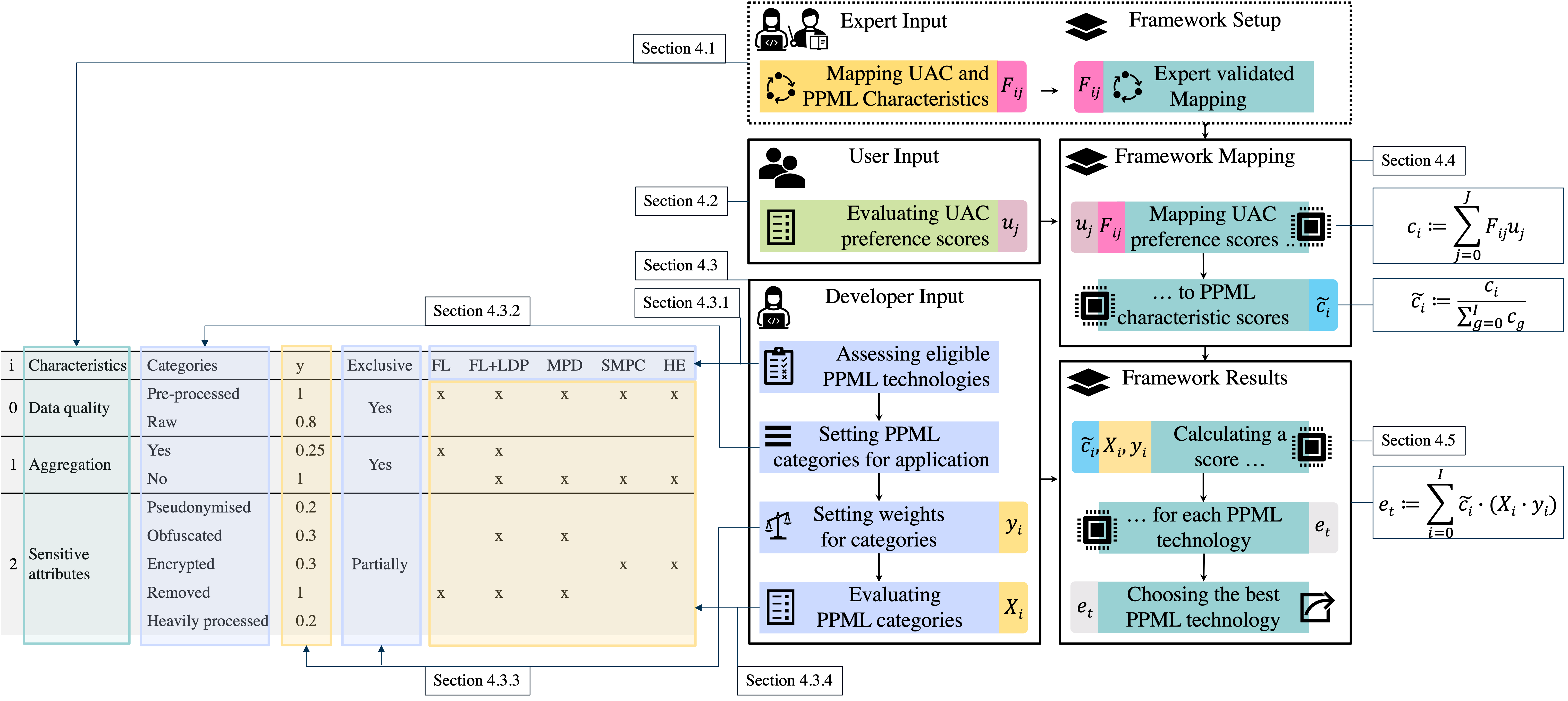}
\caption{Flowchart illustrating the procedural steps for implementing the framework. The sections of the description (section \ref{sec:framew_descr}) correspond to the framework application (section \ref{sec:application}).}
\label{fig:process}
\end{figure}

In this section, we explain the calculation steps within our framework. Figure \ref{fig:process} provides an overview of the different steps and clarifies at which state input is required to the framework. 
\par

First, the framework is set up by checking whether the mapping of UAC and PPML Characteristics is up to date and applicable to the application (see section \ref{sec:des_setup}). Second, User Input (see section \ref{sec:des_user}) and Developer Input (see section \ref{sec:des_dev}) are collected. The user input is collected by conducting a survey. We assume that the developer input is a paid service by the PPML developer. It is assessed which PPML techniques for the application are eligible (see section \ref{sec:assessing_eligible}). For each PPML Characteristics, categories (see section\ref{sec:set_cats}) with Intra-categorical weights (see section \ref{sec:setting_weights_for_cats}) are set by the developers and the PPML technologies are evaluated based on the categories (see section \ref{sec:evaluating_cats}.  
The user input is translated by applying the mapping of \citet{loebnerTB} into PPML Characteristic scores (see section \ref{sec:des_mapping}). Finally, the PPML technique scores are calculated (see section \ref{sec:des_results}). The sections of the description (section 4) correspond to the framework application (section 6). The AI service provider can now include the preference of the users for the new service into the privacy by design decisions of the application development.


\subsection{Expert Input and Framework Setup}\label{sec:des_setup}
In this step, we define the framework setup using the UAC PPML Characteristics mapping proposed by \citet{loebnerTB} (see table \ref{tab:mapping}). Their definitions of UAC are presented for the sake of completeness in table \ref{tab:UAC_definitions}. \citet{loebnerTB}, collected UAC and PPML Characteristics from existing models and grouped the attributes in several rounds of discussions with experts in privacy and machine learning. Moreover, experts validated the attributes and their influence onto each other in several feedback loops. Thus, we consider the framework setup as provided from previous work. How the proposed framework is applied to calculate the best PPML is the contribution of this paper. Thus, to later map the input of UAC to PPML Characteristics we define \( F \) as a binary mask matrix. A connection in the framework is represented by \( F_{ij} \in \{0, 1\} \), where \( F_{ij} \) denotes the connection of the \(j\)-th UAC and the \(i\)-th PPML Characteristic. The starting point are 15 UAC and 14 PPML Characteristics. For easy iteration and identification in the mapping matrix we define index \( j \) (\( J = 14 \)) and index \( i \) (\( I = 13 \)).

\begin{table}[t!]
\centering
\caption{Mapping of User Acceptance Criteria and PPML Characteristics\label{tab:mapping}}
\begin{scriptsize}

\begin{tabularx}{\linewidth}{l|ll|XXXXX:XXXXXX:XXX|}
\multicolumn{3}{l}{} & \multicolumn{13}{c}{PPML Characteristics} \\
\cline{4-17}
\multicolumn{3}{l|}{}  & \multicolumn{5}{c:}{Data} & \multicolumn{6}{c:}{Model} & \multicolumn{3}{c|}{P~\&~S}  \\
\multicolumn{3}{l|}{}  & \begin{sideways}1.~Location of data storage\end{sideways}  & \begin{sideways}2.~Data size\end{sideways} & \begin{sideways}3.~Data quality\end{sideways} & \begin{sideways}4.~Aggregation\end{sideways} & \begin{sideways}5.~Sensitive attributes\end{sideways} & \begin{sideways}1.~Explainability\end{sideways} & \begin{sideways}2.~Location of computation\end{sideways} & \begin{sideways}3.~Training method\end{sideways} & \begin{sideways}4.~Accuracy\end{sideways} & \begin{sideways}5.~Training Time\end{sideways} & \begin{sideways}6.~Performance\end{sideways} & \begin{sideways}1.~Resilience against attacks\end{sideways} & \begin{sideways}2.~Purpose and access limit. \end{sideways} & \begin{sideways}3.~Technical robustness\end{sideways}      \\ 
\cline{2-17}
& 
& PC1.~Automated decision making &     &  & \user &  &  & \user &    &  & \user & & &  &  &          \\ 
\rowcolor{gray!10} \cellcolor{white}& \cellcolor{white} & PC2.~Unauthorized secondary use~ & \both     &  & \user & \user & \both &  & \both &  &  &  &  &  & \both &          \\
 &  & PC3.~Data bias &      &  & \user &  &  & \user &  & \user &   &  &  &  &  &          \\
 \rowcolor{gray!10} \cellcolor{white}&\cellcolor{white}\multirow{-4}{*}{\rotatebox[origin=c]{90}{{\parbox[c]{0.9 cm}{Privacy concerns~}}}} & PC4.~Unauthorized access & \user     &  &  & \user & \both &  & \user &    &  \user & & & \both & \both &  \user      \\
\cdashline{2-17}
 &  & UX1.~Ease of use &  &      &  \user &  &  &  &  &    & \user & \user & \user&  &  &       \\
\rowcolor{gray!10} \cellcolor{white}& \cellcolor{white} & UX2.~Adaptability &      &  &  &  &  &  &  &  &   \user &  &  &  &  &      \\
 &  & UX3.~Availability &      &  &  &  &  &  & \user   &  &  & & \user  &  &  & \user      \\
\rowcolor{gray!10} \cellcolor{white}& \cellcolor{white} \multirow{-4}{*}{\rotatebox[origin=c]{90}{{\parbox[c]{0.9cm}{User experience~}}}} & UX4.~Performance & \user    & \user & \user &  &  &    & \user  & \user  & \user  & \user  & \user  &  &  & \\ 
\cdashline{2-17}
 &  & DP1.~Collection & \both     & \user &  & \both & \both & \user &    &  &  &  &  && \both &         \\
\rowcolor{gray!10} \cellcolor{white}& \cellcolor{white} & DP2.~Data purpose & \both     &  \user&  & \both  & \both  &  &   & \both &  &  & &  & \both &          \\
 &  & DP3.~Storage location & \both      &  &  &  & \both &  & \both &    &  &  &  &  &  &   \\
\rowcolor{gray!10} \cellcolor{white}& \cellcolor{white} \multirow{-4}{*}{\rotatebox[origin=c]{90}{\parbox[c]{0.9cm}{Perc.~data \mbox{process.}}}} & DP4.~Correctness of stored data~ & \both     &  & \both &  & \both  &  &  &  &  &  &  &  &  &     \\
\cdashline{2-17}
 &  & PT1.~Perceived lawfulness & \both     &  &  & \both & \both & \user & \both   &  &  & & & \both & \both & \user       \\
\rowcolor{gray!10} \cellcolor{white}& \cellcolor{white}  & PT2.~Fairness~     &  &  & \user &  & \both & \user &  &  \user & \user &  &&  &  & \user       \\
\multirow{-17}{*}{\rotatebox[origin=c]{90}{{User Acceptance Criteria}}} & \multirow{-3}{*}{\rotatebox[origin=c]{90}{{\parbox[c]{0.7cm}{Perc. trustw.~}}}} & PT3.~Transparency     &  &  &  &  &  & \user & \user &    &  &  &  &  &  &   \\
\cline{2-17}
\end{tabularx}\\
\hspace{5mm}P~\&~S~-~Privacy and Security\hfill \user~-~User\hspace{2mm}\both~-~User and Data Entity\\
\end{scriptsize} 
\end{table}
\begin{table}[t!]
\centering
\begin{footnotesize}
\caption{UAC and their expected influence positive (+), negative (-) or scenario dependent (*),  from \citep{loebnerTB}.}\label{tab:UAC_definitions}
\begin{tabularx}{\linewidth}{lYXl}
\hline
&UAC  & Definition & * \\ \hline 

PC1 & Automated decision making &  Concern that the process is getting out of hand and people are being treated as numbers rather than individuals \citep{smith1996information}  & -\\
PC2 & Unauthorized secondary use  & Concern of a misuse of data initially collected for a certain purpose, for a secondary purpose without authorization \citep{smith1996information} & -\\
PC3 & Data bias & Concern of discriminating results by unrepresentative data \citep{drobotowicz2021trustworthy} & -\\
PC4 & Unauthorized access & Concern of access of personal data to unauthorized people \citep{smith1996information, drobotowicz2021trustworthy} & - \\
\cdashline{1-4}
UX1 & Ease of use & Expected effort  associated with the use of a PPML technique \citep{venkatesh2012consumer, merhi2019cross}& + \\
UX2 & Adaptability & Concern that a system cannot be adapted to a change in context \citep{vierhauser2020towards} & + \\
UX3 & Availability & Level to which a user can successfully access a certain technology  \citep{villamizar2019approach} & + \\
UX4 & Performance & Extent to which the technology's use will benefit certain activities~\citep{venkatesh2012consumer, merhi2019cross} & + \\
\cdashline{1-4}
DP1 & Collection &  Concern that massive amounts of personal data are collected and stored \citep{smith1996information, zeng2020all} & *\\
DP2 & Data purpose & Has to be legitimate, explicit and specified \citep{GDPR16} & * \\
DP3 & Storage location & Transfer of data across geographical borders might reveal personal data, Article 44 ff. \citep{GDPR16}, Physical storage location can be local or with a cloud provider who has full data control and can perform malicious tasks~\citep{rao2016study} & *\\
DP4 & Correctness of stored data & Concern that the stored data exhibits errors or incorrect user data \citep{smith1996information} & * \\ 
\cdashline{1-4}
PT1 & Perceived lawfulness & Concern of a user that data is not processed lawfully \citep{GDPR16} & +  \\
PT2 & Fairness &  fairness is a substantial balancing of the involved parties to mitigate situations of unfair imbalances, where the data subject feels vulnerable. \citet{malgieri2020concept} & + \\
PT3 & Transparency & The right to receive meaningful information about the logic when automated decision-making or profiling is used \citep{GDPR16}& +  \\
\hline
\end{tabularx} 
\end{footnotesize}
\end{table}

\subsection{User Input}\label{sec:des_user}
In this step, we describe the user input. The user input contains a score for every UAC. The higher the score the more important is the UAC. We define the user input \( u \) as a vector of UAC preference scores
with \( u_j \in [0, 1] \) as the preference score for every UAC, where $ \sum_{j=0}^{J} u_j = 1 \,.$
To evaluate the preference scores $u_j$ a variety of possible methods exists and the choice for a method depends on the specific use case and thus cannot be part of this theoretical framework paper. Possible methods to conduct such a survey are e.g., Analytical Hierarchical Process (AHP) \citep{vaidya2006analytic}, Technique for Order Preference by Similarity to Ideal Solution (TOPSIS) \citep{behzadian2012state} or the Kemeny–Young method\;\citep{conitzer2006improved}.

\subsection{Developer Input}\label{sec:des_dev}
In this step, we explain how the developer's input is derived. We assume that this step is a paid service performed by framework users (developers) with expert knowledge in PPML. 

\subsubsection{Assessing eligible PPML techniques}\label{sec:assessing_eligible}
To identify which PPML techniques are taken into consideration for implementation, a list of possible methods can be found, e.g., in the international standard ISO/IEC 20889 \citep{ISOIEC20889}. This document already classifies different privacy enhancing data de-identification techniques. In addition to this list, relevant literature of most recent PPML techniques should be taken into account, to cover also new emerging techniques. 

\subsubsection{Setting PPML categories for application}\label{sec:set_cats}

A crucial step is defining categories, denoted as $k$, for each PPML Characteristic. These categories should be selected to effectively distinguish between different PPML techniques. A PPML Characteristic can have any number of categories, denoted as \(k \in [0, \infty)\). Categories can take on various forms, e.g. nominal, ordinal, or metric.

If a PPML Characteristic has only one valid category for a specific use case, we define it as a hard criterion. In such cases, if the hard criterion is not fulfilled, the technique can be directly excluded.

Let \(X_i\) be the matrix for the \(i\)-th PPML Characteristic, with dimensions \((m, n)\), where \(m\) represents the number of PPML Characteristics (\(m = I\)) and \(n\) represents the number of PPML techniques (\(n = T\)). The matrix \(X_{i,kt}\) corresponds to the \(i\)-th PPML Characteristic submatrix with \(k\) categories and $t$ PPML techniques.

\subsubsection{Setting weights for categories}\label{sec:setting_weights_for_cats}
To cover that categories can be nominal, ordinal or metric we need to weight the categories against each other. Thus we introduce a weight vector $y_i$ as the i-th y vector of matrix \( X_i \). Thereby, $y_{i,k}$ is the $k$-th category weight for the $i$-th PPML Characteristic with $y_{i,k} \in [0,1]$.
Consider the following case differentiation for $y_{i,k}$:
\[
\forall i :
\begin{cases}
    \sum_{k=0}^{K} y_{i,k} \geq 1 & \text{if exclusive}, \\
    \sum_{k=0}^{K} y_{i,k} = 1 & \text{if not exclusive}.
\end{cases}
\]

\subsubsection{Evaluating PPML categories}\label{sec:evaluating_cats}
Once the categories and weights are set up, the categories can be evaluated. The evaluation is conducted by the framework users (developers) and is expected to be performed based on previous knowledge about the respective techniques. If a technique is new, it will need to be assessed based on related literature. 
\par
If a PPML technique is assigned to a category we set \(X_{i,kt} = 1\), otherwise 0. In general, a technique can be assigned to several categories, resulting in a higher score.
\par
In the evaluation, trade-offs can be discovered. We indicate a trade-off between categories with a capital T. It is possible to set a different value below 1 making the matrix non-binary. This can be required because otherwise the result is distorted since the evaluation of the technique cannot be realised in the final implementation. The actual value for T depends on the trade-off.

\subsection{Framework Mapping}\label{sec:des_mapping}
In this step, we describe how the user input is translated into PPML Characteristics' scores. This is necessary because the user has a significant knowledge gap about PPML Characteristics and cannot evaluate them directly \citep{loebnerTB}. Thus we define \( c_i \) as the PPML Characteristic preference vector with index \( i \). It can be interpreted as the translation of UAC preference scores into preference scores for PPML Characteristics. Currently we only have the binary mask matrix $F_{ij}$. 


Next, we calculate the PPML Characteristic scores \( c_i \) that is the dot product of the binary mask matrix $F_{ij}$ and the UAC preference scores $u_j$.
\[ c_i := \sum_{j=0}^{J} F_{ij} u_j\,\,\,\,. \]

To have a better comparison of the the results, we normalise the vector \( c \). This is also required for the next calculation because these user preference scores for PPML Characteristics are later used to weight and rank the PPML techniques. Thus we calculate
\[ \forall i : \tilde{c}_i := \frac{c_i}{\sum_{g=0}^{I} c_g}\,\,\,\,. \]

\subsection{Framework Results}\label{sec:des_results}
In this step, we calculate the final ranking of PPML techniques. Once all inputs are ready, the evaluation vector \( e \) can be calculated. Each PPML technique will receive a score. If other criteria besides UAC and PPML Characteristics should be included in the ranking, this has to be taken into account by the developers.  In our framework, the PPML technique with the highest score will meet the UAC preference score the best. We calculate the evaluation vector \( e \) so that 
\[ \forall t : e_t := \sum_{i=0}^{I=13} \tilde{c_i} \cdot (X_i^{T} \cdot y_i)\,\,\,\,. \]

\section{Use Case: PSI Detection in texts}\label{sec:scenario}
To exemplify the use of the proposed framework, we describe a simplified AI application that a provider could be planning to develop. We will use the example of PSI detection. 

\subsection{Use Case description}
The application's objective is to detect the disclosure of Privacy Sensitive Information (PSI) in social media posts. The application should detect whether PSI was included in the text and classify the text accordingly. 
If the text was classified as including sensitive information, then the application would additionally detect the type of PSI contained in the text. 

\begin{figure}[tb!]
\centering
\includegraphics[width=0.5\textwidth]{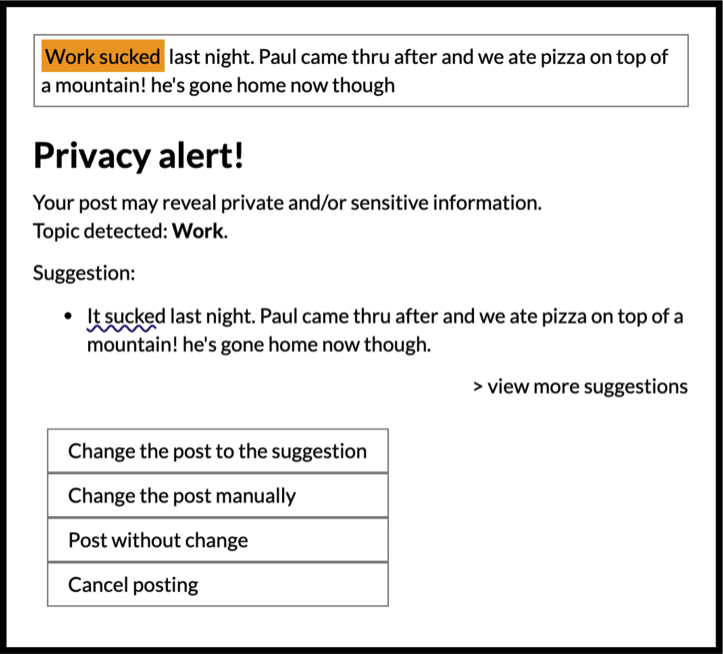}
\caption{Example of a privacy tool for PSI detection in social media texts. \citep{bracamonte2022all}}\label{fig:privacytool}
\end{figure}

Different PPML methods could be used to implement such a PSI detection application. PSI detection can be accomplished using multi-class or multi-label classification approaches that would classify private data into different categories (e.g., tracking, financial or medical data) \citep{loebnerPST}. Users' privacy concerns towards this type of privacy-preserving application have been investigated by~\citet{bracamonte2022all}, who found that users have privacy concerns with regard to perceived surveillance and perceived intrusion and secondary use, which motivates the need of PPML in this type of application. Figure \ref{fig:privacytool} shows a possible interface of such a tool.

\subsection{PSI detection requirements}\label{sec:assumptions}
Because of the stake of showcasing the framework, we have to set some implementation requirements and assumptions to narrow down the use case:
\begin{itemize}
    \item The user input that is a ranking of UAC is assumed to be already achieved by a survey. How the survey is conducted precisely is part of our future experiments.
    \item For our example, we analyse a text reinforcement learning problem for PSI detection.
    \item The data used for training contains private data that cannot be removed.
    \item A pre-trained model based on data from the data entity already exists. We do not analyse the protection during that pre-training phase.
    \item PPML should be applied only to the PSI detection of user data consisting only of the input and output data but not training data provided by the data entity.
    \item Input data is used for improving the model. This data needs also to be protected.
    \item The accuracy metrics of the PPML algorithm should be as high as possible to guarantee the privacy by design principle of full functionality.
    \item We assume the developers using our framework to be honest and unbiased.
    \item If possible we want to use BERT as ML technique in combination with the respective PPML technique because it has shown very good results for text classification tasks \citep{karl2023transformers, loebnerPST}.
\end{itemize}

\subsection{PPML for text classification}\label{sec:background}
In this section, we collect examples of text classification tasks using PPML, also highlighting some limitations and advantages. This is relevant to ensure that the PPML technique is applicable.

\par 

\textit{FL} is benchmarked by \citet{lin2021fednlp} who use the 20news dataset and 100 clients. They identify future potential for improved performance. Moreover, incorrect or biased data might cause biased decisions. They evaluate the current state of FedNLP as “relatively immature". \citet{sidhpura2022fedspam} and \citet{thapa2020fedemail} propose a SMS spam and ham classification. This task is very similar to a binary PSI detection because the length of text is comparable. \citet{liu2021federated} analyse FL for Natural Language Processing (NLP) and find relevant challenges for text classification, e.g. solving communication overhead from client updates using distillation methods. A benchmarking of FL with NLP is provided by \citet{lin2021fednlp} who analyse a combination of NLP with BERT and FL.

\textit{FL+LDP} is combined by
\citet{lobner2023enhancing} to increase the privacy of users. For a noise multiplier of 0.99 and an $\epsilon$ of 23 they achieve an F1-Score of 0.94 after 20 federated rounds for 10 clients. In this approach LDP is applied on the gradients of the users' local models and not on the text itself. The evaluation of the whole 
model took 144 min. \par

\textit{DP} for unstructured text data was investigated by \citet{klymenko2022differential}. They find that DP does not permit inference attacks themselves but creates uncertainty about the inferred data. Regarding unstructured text, this concept can run into limitations when following the strict definition that any two texts have to be considered adjacent. Thus, the application of Metric Differential Privacy (MDP) was introduced by \citet{chatzikokolakis2013broadening} for unstructured text data. \citet{carvalho2023tem} apply MDP on the IMDB review data set, using a sentiment classification model. \citet{xu2020differentially} present a differentially private Mahalanobis mechanism for text perturbation. \citet{weggenmann2018syntf} propose an automated text anonymisation approach for text mining that fulfills DP and comes with a provable plausible deniability guarantee. Classifying text into news groups, they find that their model requires a large $\epsilon$ to achieve an acceptable utility which has negative privacy consequences.
The application of LDP for unstructured text classification is possible but comes with some drawbacks such as unclear definitions and privacy guarantees. Thus we decided to analyse MDP instead of LDP. Also the privacy accuracy trade-off has to be considered carefully.

\textit{SMPC} is used by \citet{resende2022fast} for secure classification of spam and ham emails in a two-party protocol. They notice no loss in accuracy by applying SMPC with Naive Bayes and can compute the result in 21ms for an average spam SMS of 8 unigrams. They state that in their model the service provider cannot read the text and the user does not learn anything about the model, except the classification result. \citet{knott2021crypten} present the CrypTen protocol that utilises a linear layer operating on word embedding. Compared to PyTorch they noticed a reduced speed (about 2.5-3 orders of magnitude slower) with an overall inference per sample of 30ms. Comparing a two- and three-party protocol they find that the latter is slower, due to the larger number of communication rounds required by the public division protocol. \citet{reich2019privacy} use SMPC for hate-speech detection in personal text messages. They also prove that with their classification approach based on feature extraction with logistic regression and tree ensembles the users cannot learn about the model and the service provider does not learn anything about the user's message. Based on the collected examples we conclude that SMPC is applicable in our PSI detection application, even with NN. 
\par
\textit{HE} is implemented by \citet{al2020privft} who use fully encrypted data to train an encrypted model for spam and ham classification. They achieve a run time per inference of 170ms. They have no loss on prediction accuracy compared to the baseline model without a privacy layer. Moreover, they show how inference as a service can be implemented by a service provider, ensuring privacy of the data and privacy of the model.  \citet{podschwadt2020classification} present a RNN for NLP tasks without a loss in accuracy. In their example they perform a sentiment classification on movie reviews with a runtime of 5800 ms for a batch size of 32. \citet{sun2018private} implement a FHE scheme with SIMD that performs private hyperplane decision based, Naive Base and decision tree comparisons for cyphertexts. They achieve an implementation time for Naive Bayes between 49 and 142 ms for different security parameters ($\lambda$). 
Thus HE is suitable for our application but is compared to SMPC a little bit slower. Also a trade-off between security and performance exists.
\par
\textit{TEEs} are secure and isolated processing environments, distinct from the regular processing environment. By partitioning applications and running sensitive code in the TEE the security and data integrity are increased \citep{ekberg2013trusted}. \citet{sartakov2021spons} describe the use of TEE in clouds to deploy applications that are protected from e.g. unauthorised access by cloud service providers. Using lift-and-shift models whole VMs can be run in TEEs. Thus we consider this technology.

\section{Framework Application}\label{sec:application}

In this section, we apply the framework as described in section \ref{sec:framew_descr} for the PSI detection use case as presented in section \ref{sec:scenario}. In the role of framework users, we will go through each step of the framework and calculate the PPML technology that meets the user requirements the best.

\par

\subsection{PSI: Expert Input and Framework Setup}

We do not implement any changes to the framework itself and consider all UAC and PPML Characteristics to be relevant for the PPML use case. We do not include additional features. Experts should re-evaluate the framework at regular intervals based on technology changes.


\subsection{PSI: User Input}

\begin{figure}[htb!]
\centering
\includegraphics[width=0.6\textwidth]{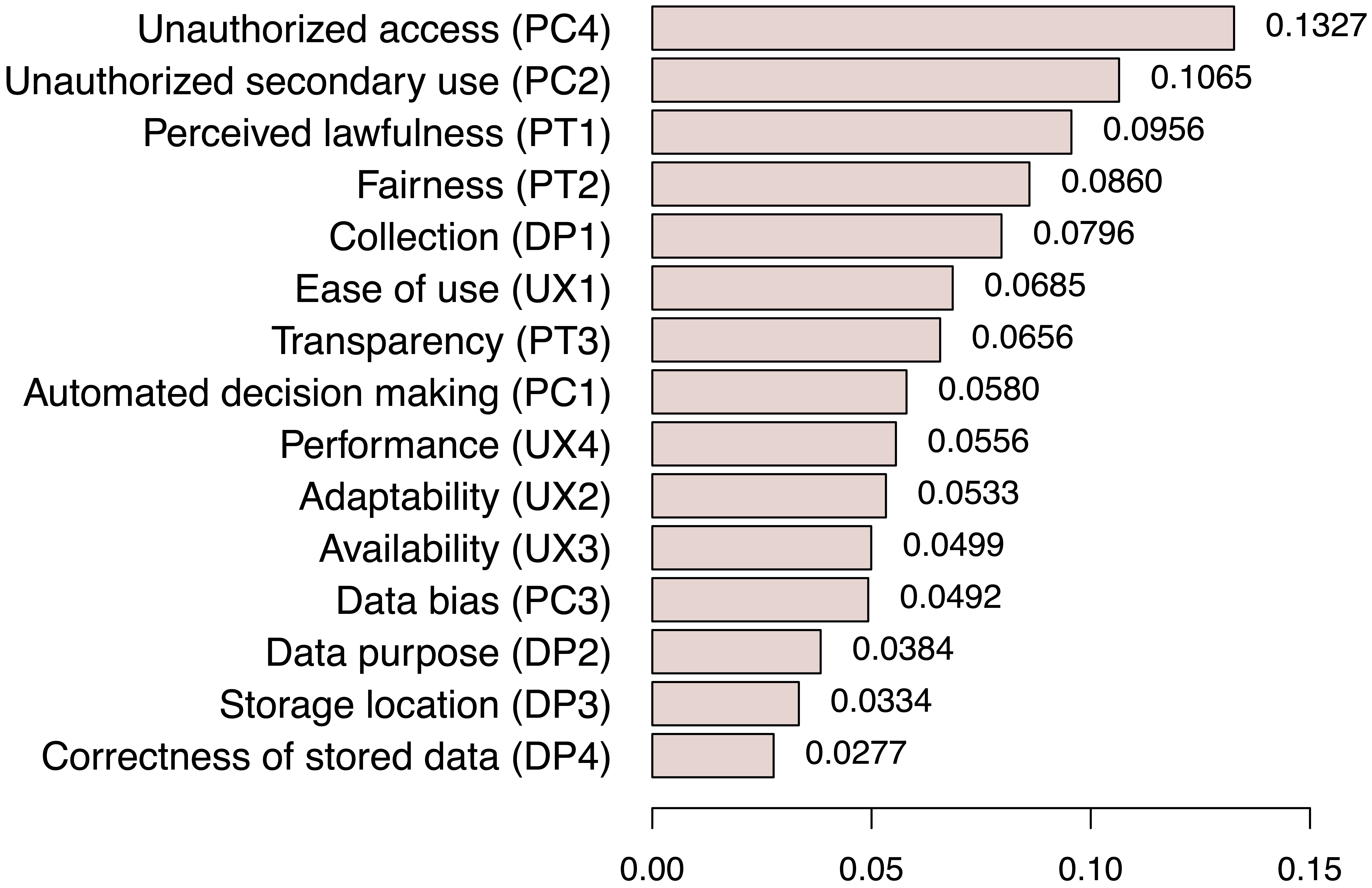}
\caption{Importance scores $u_{j}$ (x-axis) of UAC (y-axis)\label{fig:UAC_scores}}
\end{figure}

For the ranking we have run an AHP with 55 participants. Due to the high number of UAC\;(15), the UAC were divided into four groups (PC, UX, DP, PT) and the respective subgroups as provided in table \ref{tab:mapping}. The AHP took each participant around 15 minutes and a tolerable \citep{hummel2014group} Consistency Ratio\;(CR) below 0.2 was achieved for 31-33 participants in each subgroup. Participants above 0.2 were excluded. Between 18 and 21 participants achieved a CR score below 0.1. 27 participants were male and 27 female, 1 participant was diverse. Four participants were in the age group 18-19, 42 between 20-29, seven between 30-39, none between 40-49, one between 50-59 and one between 60-69. We aimed specifically for a younger age to get people with high social media interaction. The usage and posting behavior is shown in table \ref{tab:demographics}. Following \citep{saaty2004decision} when using AHP for group decisions we expect a relative importance score for each feature between 1 and 0. All global feature preference values sum up to 1. We also follow the fundamental scale: \{extremely important : 9, very strong important : 7, strong important 5, moderate important : 3, equal : 1 \}. We have also included intermediate choices (2, 4, 6, 8). The global preference values are provided in figure \ref{fig:UAC_scores}.
\begin{table}[t!]
\centering
\begin{footnotesize}
\begin{tabularx}{.85\textwidth}{lXX}
\hline
Quantity & How often do you use social media? & How often do you post in social media? \\ \hline
Daily      & 33 & 4  \\
Weekly   & 5  & 11 \\
Monthly     & 11 & 22 \\
Yearly      & 1  & 12 \\
Not at all      & 5  & 6  \\ \hline
\end{tabularx}
\caption{Social media usage and posting behavior}
\label{tab:demographics}
\end{footnotesize}
\end{table}

\subsection{PSI: Developer Input}
In this section, we describe how we derived the developer input for the PSI detection application.

\subsubsection{PSI: Assessing eligible PPML techniques}
Techniques are selected based on related literature, ISO/IEC 20889 and our own experience. We assume that also developers would choose techniques based on their own experiences and the techniques establishment. As shown in section \ref{sec:background} all techniques chosen for showcasing are suitable for the PSI detection application:
\begin{itemize}
    \item Federated Learning (FL)
    \item FL + Local Differential Privacy (FL+LDP)
    \item Metric Differential Privacy (MDP)
    \item Secure Multiparty Computation (SMPC)
    \item Homomorphic Encryption (HE)
    \item Central Trusted Execution Environment (TEE)
\end{itemize}

\subsubsection{PSI: Setting PPML categories for application}\label{sec:categorical_scales}
In this section, we elicit categories for the PPML Characteristics based on table \ref{tab:mapping}. An initial definition of the PPML Characteristics is already provided by \citet{loebnerTB}. We have separated the PPML Characteristics into categorical attributes that do not have a logical order, but one category might be preferred by the users. 
\par
For a better overview and easier computation of the final recommendation, we start by defining hard requirements that will lead to an exclusion of the PPML if not fulfilled. We do this for an improved efficiency of the evaluation process. The following criteria of PPML Characteristics we define as a hard requirement.

\textit{Location of raw data storage:} The location where raw user data is stored is for our PPML techniques in focus local (stored on a user's device), central (stored on a central server within the AI service chain) or distributed (stored across several entities). For our PSI application we set the requirement to only allow locally stored raw user data. We set this restriction because survey participants clearly preferred raw data to be stored exclusively local.

\textit{Data size:} Since for some ML models, a huge data size is mandatory, a threshold should be set for the data size that indicates the minimum amount of data, e.g. in X (formally known as Twitter) posts, required to achieve a useful accuracy. In the papers investigated by \citet{loebnerPST}  the amount of text samples in the data set was between 1,000 and 800,000. Thus, we will just provide the threshold of minimum 1,000 samples to indicate that for the application pre-trained models have to be used. Since we only consider short text posts on social media, we neglect the text-length. 
\par
\textit{Training method:} Not all PPML techniques are suitable for each training method. Frequently used training methods are e.g., \textit{supervised learning, semi-supervised learning, unsupervised learning} or \textit{reinforcement learning} \citep{loebnerTB}. Thus we will evaluate for common tasks whether our PPMLs in focus for the PSI application are suitable for a NLP task that is part of reinforcement learning. 
\par
Next, we evaluate the hard requirements for all PPML Characteristics (see table \ref{tab:hardrq}). 

Regarding the \textit{location of data storage}, all technologies except a central TEE allow to store raw data exclusively local. We characterise raw data as the unaltered user's text that contains potential private information. FL, FL+LDP, SMPC and HE alter the data locally, e.g. on a mobile phone before sharing with other entities \citep{lin2021fednlp, lobner2023enhancing, carvalho2023tem, resende2022fast, al2020privft}. With a central TEE the data is required to be send to a trusted entity running the TEE. The PPML Characteristics \textit{data size} and \textit{training method} are fulfilled by all PPML technologies since comparable application were found in section \ref{sec:scenario}.  
\begin{table}[t!]
\centering
\caption{Hard requirements for exclusion of technologies including PPML Characteristics and categories (see section \ref{sec:categorical_scales}), and PPML technique evaluation (see section \ref{sec:psi_eval}). \label{tab:hardrq}}
\centering
\begin{footnotesize}
\newcommand{\xstar}{\makebox[0pt][l]{x*}\phantom{x}}
\begin{tabularx}{.8\linewidth}{lXcccccc}
\hline
\parbox[t]{2cm}{\textbf{Characteris.}}                           & \textbf{Categories}               & \textbf{FL} & \textbf{FL+LDP }& \textbf{MDP} &\textbf{SMPC} & \textbf{HE} & \textbf{TEE}\\ \hline 

\multirow{3}{*}{\parbox[t]{2cm}{Location\;of raw\;data storage}}  & Local                    & x  & x     & x  & x    & x &  \\
                                               & Distributed              &    &      &    &      &   & \\
                                               & Central                  &    &      &    &      &    &x\\  \hline

Data size                                      & Min. 1\,k sampl.     & x  & x     & x  & x    & x  &x\\  \hline
 \multirow{4}{*}{\parbox[t]{2cm}{Training method}}               & Supervised     &    &       &    &   &   &    \\
                                               & Semi-supervised  &    &       &    &      &   & \\
                                               & Unsupervised     &    &       &    &      &   & \\
                                               & Reinforcement    & x  & x     & x& x    & x & x \\

\hline
\end{tabularx}
\end{footnotesize}

\end{table} 
\par

Next, we explain how we set up the categorical scales for the different PPML Characteristics that are not hard requirements. All PPML Characteristics that are not hard requirements are provided with their respective categories in table \ref{tab:y_exclusive}. It is important to keep in mind that for other applications and especially other PPML techniques, the scales might require an extension.

\textit{Data quality:} First, data quality is influenced by outliers and missing data that cannot be handled well by all algorithms. Second, data quality is influenced by the process of removing, aggregating, obfuscating, or changing data for de-identification. Thus there is an inverse relation with the attribute \textit{aggregation}. Regarding text data for PSI detection, it is known that the data has to be pre-processed since texts from social media are full of shortened words, abbreviations or colloquial language that complicate the analysis \citep{loebnerPST}.   \par

\textit{Aggregation:} According to ISO/IEC 20889 aggregated data is representing statistically the attributes of several data entities and is only useful in specific use cases. Aggregation has an inverse relation with \textit{data quality} and \textit{accuracy}. We indicate with \textit{yes} or \textit{no} whether \textit{aggregation} is happening on the raw data during the de-identification. \par

\textit{Sensitive attributes:} Plain sensitive data is one of the most important issues in PPML. To prevent the revealing of plain sensitive data we differentiate between the following fundamental methods of \textit{pseudonymisation, obfuscation, encryption, removal} and \textit{heavy processing}. \textit{Heavily processed data} we define as processed in a way that the original private data is no longer recognisable. Although removing sensitive data is the best method to break re-identification attacks, this can be rarely done if accuracy is required to be high. \par

\textit{Explainability:} Explainable results help to deploy proper PPML techniques \citep{xu2021privacy}. There exist \textit{ante-hoc} methods that are explainable by default such as decision trees, \textit{post-hoc} methods that are methods applied after the model was built  \citep{carvalho2019machine}, \textit{global methods} that explain the overall general behaviour of the model \citep{burkart2021survey} and \textit{local methods} that aim to explain individual, specific decisions  \citep{adadi2018peeking}. 
\par
\textit{Location of computation:}
For simplicity reasons, we distinguish between three possible locations of computation. These are \textit{cloud} computation, computation on \textit{local} user devices and \textit{distributed} computation. Computation refers to the training of the model \citep{loebnerTB}.   

\par
\textit{Accuracy:} An accuracy threshold for the suitable accuracy metrics should be defined. If a model in combination with the PPML is not able to achieve the threshold, the PPML should not be further considered. From \citet{loebnerPST} we know that for a binary classification of PSI data into privacy sensitive and non-sensitive texts, related models achieve an F1-score of 0.98 \citep{kopeykina2020photo}. Thus, for the binary classification we decided for the categories $< 0.84, 0.85-0.89, 0.9-0.94, 0.95-1.00$.
\par
\textit{Training time:} 
Training time influences the adaptability of a PPML model and depends on the computational complexity of the training task \citep{loebnerTB}. Especially for reinforcement learning problems large amounts of training time are required which are often compensated with the use of GPUs or FPGAs \citep{rothmann2022survey}. Training time is very hard to compare between different approaches because complexity of models, data sets, computational resources and parties involved differ. Nevertheless, we want to elicit from existing models first insights on the suitability for PSI detection. To cover this issue we set a very general scale of categories  $<12$h, 12h-24h, $>24$h.
\par
\textit{Performance:} 
Performance is the overall runtime of an application. Regarding the PSI classification task, it is the time between user request and receiving the classification result. \citet{egger2012waiting} elicited three thresholds from literature. First, 0.1\;s is the threshold for feeling an instant reaction of the system. Second, 1.0\;s is the threshold a delay is noticed but the flow of a user's thoughts is not interrupted. Third, 10\;s is the threshold of keeping the attention on the dialogue between application and user. We follow these three limits for the evaluation of PPML techniques. 
\par
\textit{Resilience against attacks:} As explained in section~\ref{sec:threat_model}, we consider the definition of “globally private'' which refers to attackers external to the AI service chain. We consider only attacks which leak information to the external adversary, i.\,e.\ \textit{model inversion attacks}~\cite{fredrikson2015model} and \textit{membership inference attack}~\cite{shokri2017membership}. Our characteristic reflects if the considered PPML approach is resilient against this kind of attack. There also other attacks such as poisoning attacks~\cite{biggio2012poisoning} or evasion attacks~\cite{biggio2013evasion} which do not aim to infer data but to manipulate the results of the model's answer. Since these attacks are more related to the underlying ML approach we consider them under model robustness. 
\par
\textit{Purpose and access limitation:} As already explained in section~\ref{sec:threat_model}, for this characteristic we consider the definition of “locally private'' which refers to attackers inside the AI service chain. Based on \citet{tanuwidjaja2020privacy}, we differentiate between privacy of user, privacy of model and privacy of result. Due to our distinction between users of the system and data entities who only provide training data for the system, we expanded their classification by adding data entities. Contrary to the external attackers from the previous characteristic, each member of the AI service chain needs “some access'' to the data in order to fulfil their task. Therefore, for this characteristic, we evaluate whether the PPML methods implements countermeasures to protect each of these data layers to ensure members of the AI service chain do not use the data for other purposes than intended. E.g., blackbox reconstruction attacks are considered here for entities that can access the results. Again, we only focus on how the PPML method can help to improve the privacy, thus concepts such as authorisation management are not considered. 
\par
\textit{Technical robustness:} According to the Ethic Guidelines for Trustworthy AI \citep{doi/10.2759/177365} technical robustness should ensure that even with small changes in the operating environment or the context of the model the PPML behaves reliable during the whole AI life cycle, minimising unexpected or unintentional behaviour. In this framework we focus on the evaluation of the susceptibility to adversarial attacks based on existing literature, especially model poisoning \citep{chen2017targeted}. In model poisoning attacks the adversary tries to change the model's prediction by polluting the model with manipulated input data. Thereby, the attack can be targeted to achieve a certain classification result or untargeted, just polluting the model and lowering its accuracy \citep{tocchetti2022ai}. 

\begin{table}[t!]
\centering
\caption{PPML Characteristics with categories (see section \ref{sec:categorical_scales}) and category weights $y$ (see section \ref{sec:weights}).
\label{tab:y_exclusive}}
\begin{footnotesize}
\newcommand{\xstar}{\makebox[0pt][l]{x*}\phantom{x}}
\begin{tabularx}{.75\linewidth}{rlXll>{\raggedleft\arraybackslash}p{1.5cm}}
\toprule
\textbf{i} & \textbf{Characteristics} & \textbf{Categories} & \textbf{y} & \textbf{Exclusive} \\
\hline 
\multirow{2}{*}{0} & \multirow{2}{*}{Data quality} & Pre-processed & 1 & \multirow{2}{*}{Yes} \\
 & & Raw & 0.8 & \\ 
\hline
\multirow{2}{*}{1} & \multirow{2}{*}{Aggregation} & Yes & 0.25 & \multirow{2}{*}{Yes} \\
 & & No & 1 & \\ 
\hline
\multirow{5}{*}{2} & \multirow{5}{*}{\parbox[t]{2cm}{Sensitive\\ attributes}} & Pseudonymised & 0.2 & \multirow{5}{*}{Partially} \\
 & & Obfuscated & 0.3 & \\
 & & Encrypted & 0.3 & \\
 & & Removed & 1 & \\
 & & Heavily processed & 0.2 & \\ 
\hline
\multirow{4}{*}{3} & \multirow{4}{*}{Explainability} & Ante-hoc methods & 0.25 & \multirow{4}{*}{No} \\
 & & Post-hoc methods & 0.25 & \\
 & & Global explainability & 0.25 & \\
 & & Local explainability & 0.25 & \\ 
\hline
\multirow{3}{*}{4} & \multirow{3}{*}{\parbox[t]{2cm}{Location of computation}} & Local & 0.5 & \multirow{3}{*}{No} \\
 & & Distributed & 0.3 & \\
 & & Central & 0.2 & \\ 
\hline
\multirow{4}{*}{5} & \multirow{4}{*}{\parbox[t]{2cm}{Accuracy (F1-score)}} & \textless 0.84 & 0 & \multirow{4}{*}{Yes} \\
 & & 0.85-0.89 & 0.3 & \\
 & & 0.9-0.94 & 0.6 & \\
 & & 0.95-1 & 1 & \\ 
\hline
\multirow{3}{*}{6} &  \multirow{3}{*}{Training time} & $>$24h & 0.3 & \multirow{3}{*}{Yes} \\
 & & 12-24h & 0.6 & \\
 & & $<$12h & 1 & \\ 
\hline
\multirow{4}{*}{7} &  \multirow{4}{*}{Performance} & Attention lost: \textgreater 10 s & 0 & \multirow{4}{*}{Yes} \\
 & & Attention kept: 1.0-10 s & 0.3 & \\
 & & Uninterrupted: 0.1-1.0 s & 0.6 & \\
 & & Instant reaction: \textless 0.1 s & 1 & \\ 
\hline
\multirow{2}{*}{8} & \multirow{2}{*}{\parbox[t]{2.5cm}{Resilience against attacks}} & Model inversion & 0.5 & \multirow{2}{*}{No} \\
 & & Membership inference & 0.5 & \\ 
\hline
\multirow{4}{*}{9} & \multirow{4}{*}{\parbox[t]{2cm}{Purpose and access limitation}} & Privacy of data entity & 0.25 & \multirow{4}{*}{No} \\
 & & Privacy of user & 0.25 & \\
 & & Privacy of model & 0.25 & \\
 & & Privacy of result & 0.25 & \\ 
\hline
\multirow{2}{*}{10} & \multirow{2}{*}{\parbox[t]{2.5cm}{Technical\\robustness}} & Poisoning countermeasures & 1 & \multirow{2}{*}{No} \\
 & & No countermeasures & 0 & \\ 
\bottomrule
\end{tabularx}
\end{footnotesize}
\end{table}

\subsubsection{PSI: Setting weights for categories}\label{sec:weights}
For each category matrix $X_i$ we have to set a weight vector $y_i$. The sake of these weights is to support framework users with another dimension to formulate requirements. Again the sum of each weight vector if not exclusive has to add up to 1. Setting up the weight vector is a task that is performed in the AI service chain by the party who has the most knowledge about the PPML characteristics. To achieve an equal scaling, an AHP could be used again for each category. All values for $y$ are shown in table \ref{tab:y_exclusive} where we also indicate whether a category is exclusive or not. If a category is not exclusive ("No"), then $\sum{y} = 1$ holds.
For the sake of showcasing, we will set the weight vectors exemplary based on authors' experience (see table \ref{table:y_justification}) 

\begin{table}[!t]
\centering
\caption{Attributes, Justifications, and \(y\) Values}
\label{table:y_justification}
\begin{footnotesize}
\begin{tabularx}{\linewidth}{lXl}
\hline
\textbf{PPML Characteristic} & \textbf{Justification} & \textbf{\(y\)} \\
\hline
Data Quality & The effort to create pre-processed text data from raw is low because a lot of packages exist already and the BERT model has its own pre-processing layer. & \([1, 0.8]\) \\

Aggregation & Can cause information loss, thus no aggregation is preferred. & \([0.25, 1]\) \\

Sensitive Attributes & Handled differently, with the removal having the highest protection, and obfuscation and encryption evaluated as effective but vulnerable to attacks. & \([0.2, 0.3, 0.3, 1, 0.2]\) \\

Explainability & The more explainability the better. & \([0.25, 0.25, 0.25, 0.25]\) \\

Location of Computation & Data not leaving the device cannot be sold to third parties or accidentally altered. This also holds for distributed setups although the privacy guarantees are often lower compared to local computation. & \([0.5, 0.3, 0.2]\) \\

Accuracy & The higher the better but below 0.84 is not acceptable. & \([0, 0.3, 0.6, 1]\) \\

Training Time & Over 24h quick adaptation to environmental changes is prevented. & \([0.3, 0.6, 1]\) \\

Performance & Faster is better but losing attention is not acceptable. & \([0, 0.3, 0.6, 1]\) \\

Res. Against Attacks & All attacks should be prevented. & \([0.5, 0.5]\) \\

Purpose and Access Lim. & Should be equally implemented. & \([0.25, 0.25, 0.25, 0.25]\) \\

Technical Robustness & Important over the whole PPML life cycle; data poisoning needs to be prevented. & \([1, 0]\) \\
\hline
\end{tabularx}
\end{footnotesize}
\end{table}

\subsubsection{PSI: Evaluating PPML categories}\label{sec:psi_eval}

In this section we describe the evaluation of PPML technologies (FL, FL+LDP, MDP, SMPC, HE) by labelling them into PPML categories from section \ref{sec:categorical_scales}. For the sake of showcasing, the authors take over the role of framework users. To evaluate the technologies we use the first 50 papers in ACM, IEEE and Google Scholar that were identified by search terms that consist of the respective technology, PPML Characteristic respective PPML criteria. The results of the evaluation are presented in table \ref{tab:PPML_metrics}. 
\par
\textit{Data quality:} To achieve a high accuracy in the PSI detection task, pre-processing is required for NLP. All PPML are compatible with our requirement to work with BERT \citep{basu2021benchmarking, akimoto2023privformer, lee2022privacy} that requires pre-processing. The pre-processing usually happens in a pre-processing layer of the BERT model.  
\par
\textit{Aggregation:} A typical task within FL is to aggregate the gradients of the clients with an aggregation function to create an updated model \citep{mothukuri2021survey}. Thus aggregation is automatically happening for FL and FL+DP. The updated model might, due to aggregation, not contain the best gradients for a specific client. MDP, SMPC and HE are not using aggregation before the actual ML task. 
\par
\textit{Sensitive attributes:} FL is preserving privacy by the “lessening footprint'' of private user data in the gradients that are submitted to the central server \citep{mothukuri2021survey}. Thus we introduced the category \textit{heavily processed} because in contrast to \textit{removed} the private attributes can be reconstructed through several attacks, e.g. by a malicious central server. Regarding FL+LDP, an additional layer of privacy is added to the gradients before they are shared with the central server. In the ISO/IEC 20889 LDP is evaluated as useful for data minimisation and in scenarios where the data receiving entities cannot be trusted. LDP uses random noise from a “carefully selected'' probability distribution that is added to the data to make the algorithm differentially private while simultaneously preserving the desired usefulness \citep{ISOIEC20889}, thus we map this as \textit{obfuscation}. We follow \citet{tanuwidjaja2020privacy} and classify SMPC and HE as cryptographic approaches that use encryption. HE is computing on encrypted data thus the main technique here is \textit{encryption}. 
\par
\textit{Explainability:} In FL, it is possible to use all existing explainability-methods but it is important to differentiate between the explainability of the local and global model with a focus on not revealing private data. Regarding DP all explainability methods are possible \citet{patel2022model, naidu2021differential}. Thus, the same holds for FL+LDP. Regarding explainability in SMPC and HE we did not find reliable results. Therefore, we assume that black-box methods will work and post-hoc methods and local methods are possible. Due to the encryption we determine that ante-hoc methods and global explainability are not suitable or hard to implement. These methods could reveal too much information about the model and thus leverage attacks.
\par
\textit{Location of computation:} FL as used in our scenario relies similar to \citet{lin2021fednlp} on a central server. Computation is done mainly locally but for the model updates a central model is computed with the clients' gradients. This improved model is sent back to the clients as an updated model. Regarding FL+LDP the \textit{locations of computation} does not change since with LDP only an additional step on the client is added that is also computed locally \citep{feyisetan2019leveraging}. In contrast to LDP, we assume MDP to be applied centrally before data is computed. SMPC uses distributed protocols to derive a result, thus the computation is happening distributed \citep{resende2022fast}. HE computes results and model centrally \citep{al2020privft}.
\par
\textit{Accuracy (F1-Score)}: For the problem of PSI detection we expect for a binary text classification a F1-Score of 0.97 or better \citep{kopeykina2020photo}. Since the referred models in this section do not deal with the same data and classification problem, accuracy scores and metrics provide just a range of expected accuracy. Related work achieved for FL in a binary text classification task an F1-Score over 0.95~\citep{thapa2020fedemail, sidhpura2022fedspam}. Text classification tasks with FL and an additional LDP layer exhibit a reduced accuracy due to the privacy-utility trade-off, achieving a F1-Score between 0.85 and 0.89 \citep{lobner2023enhancing} for a binary text classification task. Also \citet{basu2021privacy} notice a reduction of accuracy caused by DP in comparison to their baseline model of approximately 5\%. \citet{carvalho2023tem} achieve an average accuracy of 75\% using an MDP approach. SMPC in general does not exhibit the issue of accuracy loss and thus an accuracy over 0.95 is expected~\citep{resende2022fast}. Also HE has no expected loss of accuracy compared to a baseline model trained without PPML, thus we expect an F1-Score over 0.95~\citep{lee2022privacy}.
\par
\textit{Training time}: \citet{xu2020research} report a training time for PSI detection in texts for LSTM of approximately 16\;h and CNN off approximately 2\;h. For FL in tests with 10 clients for a binary text classification task of spam and ham, 40 training rounds with an LSTM model took \citet{lobner2023enhancing} 137 minutes. Adding an LDP layer for 10 clients extended the training time to 144 minutes. Regarding HE studies report a training time of 5.04 days \citep{al2020privft}. One reason identified a huge data transfer rate between GPU and CPU. Regarding MDP we assume that there is no notable delay caused to the training time. Although we did not identify any related work reporting training times for MDP in text classification, the training performance itself is expected to be increased only slightly because more epochs might be required. To the best of our knowledge, no reports about training time for privacy preserving training exist. Most studies identified \citep{lu2023bumblebee, resende2022fast, wang2022mpc, li4453303priber} rely on pre-trained models but do not train with secure multiparty computation itself. Thus we do not tick a box for SMPC which will result in a lower score compared to the other technologies.


\textit{Performance}: \citet{xu2020research} report for their best model a detection time of 0.21\;s what we denote as uninterrupted classification. Since FL and FL+LDP rely on local models, we expect a time similar to the baseline model and thus, classify the performance as uninterrupted. \citet{resende2022fast} evaluate the runtime of their SMPC classification model in the worst case 0.3\;s and in the best case 0.022\;s and outperform \citet{reich2019privacy} who report a total classification time between 7.2 and 13.3\;s. Also \citet{knott2021crypten} report a classification of text in 0.03\;s. In average we assume an uninterrupted performance for SMPC. \citet{al2020privft} report the computation of a classification result within 0.3\;s. Regarding HE we identified a prototype reporting a performance of approximately 6\;s and thus, set \textit{attention kept} \citep{podschwadt2020classification}. For MDP we have not found any reports about the performance. Based on LDP performance \citep{fan2020privacy} and the short text-length, we expect the performance to be uninterrupted.

\textit{Resilience against attacks:} FL alone is not resilient against model inversion, adversarial attacks or membership inference attacks \citep{mothukuri2021survey}. Adding an LDP layer, the resilience against external adversaries can be improved \citep{lobner2023enhancing}, protecting the model from membership inference attacks. There is a trade-off between accuracy and privacy that needs to be considered. There is no sufficient protection against model inversion attacks because updated models are not protected. Also MPD is applied on the data and thus helps only with membership inference but not with model inversion attacks. Regarding SMPC, the model and the users' data are well protected against external adversaries \citep{resende2022fast}. Also HE has the potential to protect against model inversion and membership inference attacks from external adversaries. In their model that combines BERT and HE \citet{lee2022privacy} significantly reduce the inversion risk, especially the black box inversion attacks cannot be applied using an 128-bit security level. With only the client being able to decrypt the classification result, membership inference attacks from external adversaries can be prohibited. 

\textit{Purpose and access limitation:} \citet{kairouz2021advances} find that although FL improves privacy compared to centralized training approaches no formal privacy guarantees exist. Thus information leakage between client and central server is possible. Especially over several rounds of training the central server can learn model parameters and compromise user data. Since the data from the data entity is normally used to pre-train a model by the central server, there is no protection against attackers from inside at all. Adding a differential privacy layer on top of each client helps to protect the local model. Thus privacy of model and privacy of user are increased but a trade-off exists \citep{lobner2023enhancing}. On the one hand, it takes more federated rounds until the central server can learn about the model, on the other hand, the DP layer is very likely to reduce accuracy of the model. The DP layer at the client does not protect the data entity. MPD can be applied on the user data as well as on the data from the data entities, but it does not help to protect the model or results. Still, a privacy-accuracy trade-off exists. Regarding SMPC other participants are learning nothing about the result or the user data \citep{resende2022fast}. We assume that the protocol is protected from user access and thus privacy of the model can be achieved. \citet{al2020privft} show that HE can protect the input data (privacy of data entity, privacy of user), privacy of model and the classification result because each client encrypts the data before passing it to the central server \citep{tanuwidjaja2020privacy}.   

\textit{Technical robustness:} In this section, we evaluate whether PPML techniques are vulnerable against model poisoning and whether mitigation strategies already exist. 
Model poisoning is a huge problem in FL \citep{cao2019understanding} but countermeasures exist already for FL that guarantee comparable performance \citep{liu2021privacy, ma2022shieldfl}. Regarding LDP protocols are in theory vulnerable to model poisoning \citep{bohler2021secure, xu2021privacy}, e.g. \citet{cao2021data} show that it is possible for an adversary to inject fake users to an LDP protocol and successfully send carefully manipulated data that is accepted by the central server. \citet{xu2021mitigating} analyse mitigation strategies for poisoning in DP in text classification problems and reduce the attack success rate from 0.94 to 0.008 while accuracy is only reduced by 0.005. Regarding SMPC \citep{chaudhari2022safenet} and HE we expect that model poisoning is also possible and needs countermeasures if the user data is also used for training. We identified a lack in literature investigation model poisoning in SMPC and HE.

 \begin{table}[htbp!]
 \centering
\caption{PPML Characteristic evaluation (see section \ref{sec:categorical_scales}). Evaluation (see section \ref{sec:psi_eval})\label{tab:PPML_metrics}}
\begin{footnotesize}
\newcommand{\xstar}{\makebox[0pt][l]{x*}\phantom{x}}
\begin{tabularx}{.8\linewidth}{lXlllll}
\hline
\parbox[t]{2cm}{\textbf{Characteristics}}                           & \textbf{Categories}               & \textbf{FL} & \textbf{FL+LDP }& \textbf{MDP} &\textbf{SMPC} & \textbf{HE} \\ \hline 
\multirow{2}{*}{Data quality}                  & Pre-processed            & x  & x     & x  & x    & x  \\
                                               & Raw                      &    &       &    &      &    \\ \hline
\multirow{2}{*}{Aggregation}                   & Yes                      & x & x     &    &      &    \\
                                               & No                       &    &       &  x  & x    & x  \\ \hline
\multirow{5}{*}{\parbox[t]{2cm}{Sensitive\\ attributes}}        
                                               & Pseudonymised            &    &       &    &      &    \\
                                               & Obfuscated               &    & x     & x  &      &    \\
                                               & Encrypted                &    &       &    &  x   & x  \\
                                               & Removed                  &    &       &    &      &    \\ 
                                               & Heavily processed        &  x &    x  & x   &      &    \\        \hline
\multirow{4}{*}{Explainability}                & Ante-hoc methods         &  x  &  x     & x  &      &    \\
                                               & Post-hoc methods         & x  & x     &  x  &   (x)   &  (x)  \\
                                               & Global explainability    & x   &  x     & x  &      &    \\
                                               & Local explainability     & x  &   x    &  x  &  (x)    &  (x)  \\ \hline
\multirow{3}{*}{\parbox[t]{2cm}{Location\;of computation}}       & Local                    & x  & x     &    &      &    \\
                                               & Distributed              &    &       &    & x    &    \\
                                               & Central                  & x  & x     & x  &      & x  \\ \hline
\multirow{4}{*}{\parbox[t]{2cm}{Accuracy (F1-score)}}           & \textless 0.84           &    &       & T &      &    \\
                                               & 0.85-0.89                &    & T     &    &      &    \\
                                               & 0.9-0.94                 &    &       &    &      &    \\
                                               & 0.95-1                   & x  &       &    & x    & x  \\ \hline
\multirow{2}{*}{Training time}                 & $>$24\,h                &     &       &    &   (x)   &    x \\
                                               & 12-24\,h            &   &       &    &      &    \\ 
                                               & $<$12\,h            &  x &  x     &  (x)  &      &    \\ \hline
                                               & Attention lost: \textgreater 10\,s      &    &       &    &      &    \\
\multirow{3}{*}{Performance}                   & Attention kept: 1.0-10\,s      &    &       &    &      &  x  \\
                                               & Uninterrupted: 0.1-1.0\,s              &   x &   x    &  (x)  &   x   &    \\
                                               & Instant reaction: \textless 0.1\,s          &   &      &   &      &    \\ \hline
\multirow{2}{*}{\parbox[t]{2.5cm}{Resilience against attacks}}    & Model inversion          &    &       &    & x    & x  \\
                                               & Membership inference     &    & T     & T  & x    & x  \\ \hline
\multirow{3}{*}{\parbox[t]{2cm}{Purpose and\;access limitation}} 
                                               & Privacy of data entity   &    &       & T  &      & x  \\
                                               & Privacy of user          &    & T     & T  & x    & x  \\
                                               & Privacy of model         &    & T      &    &  (x)    & x  \\
                                               & Privacy of result        &    &      &    & x    & x  \\ \hline
\multirow{2}{*}{\parbox[t]{2.5cm}{Technical\\robustness}}    
                                               &  Poisoning counterm.               &  x  &  x     & x  &      &    \\
                                               &  No countermeasures         &    &       &    &    (x)  & (x)  \\ 
                                               \hline
                                               \multicolumn{6}{l}{\scriptsize{Based on literature := x,  Trade-off := T, Estimate := (x)}}\\

\end{tabularx}
\end{footnotesize}
\end{table}


\subsection{Framework Mapping}
In this section we describe how to map the importance scores of the PPML characteristics and the PPML category evaluation metrics. As mentioned in the assumptions we only take the user and not the data entity into account.
The computed normalised $\tilde{c}_i$ values for each PPML characteristic are provided in figure \ref{fig:UAC_scores} for User and Data entity. The translation of UAC preferences scores into PPML Characteristic preference scores is presented in figure \ref{fig:scores}.

\subsection{PSI: Framework Results}\label{sec:psi_fram_res}
\begin{figure}[htb!]
\centering
     \begin{subfigure}[b]{0.48\textwidth}
         \centering
         \includegraphics[width=\textwidth]{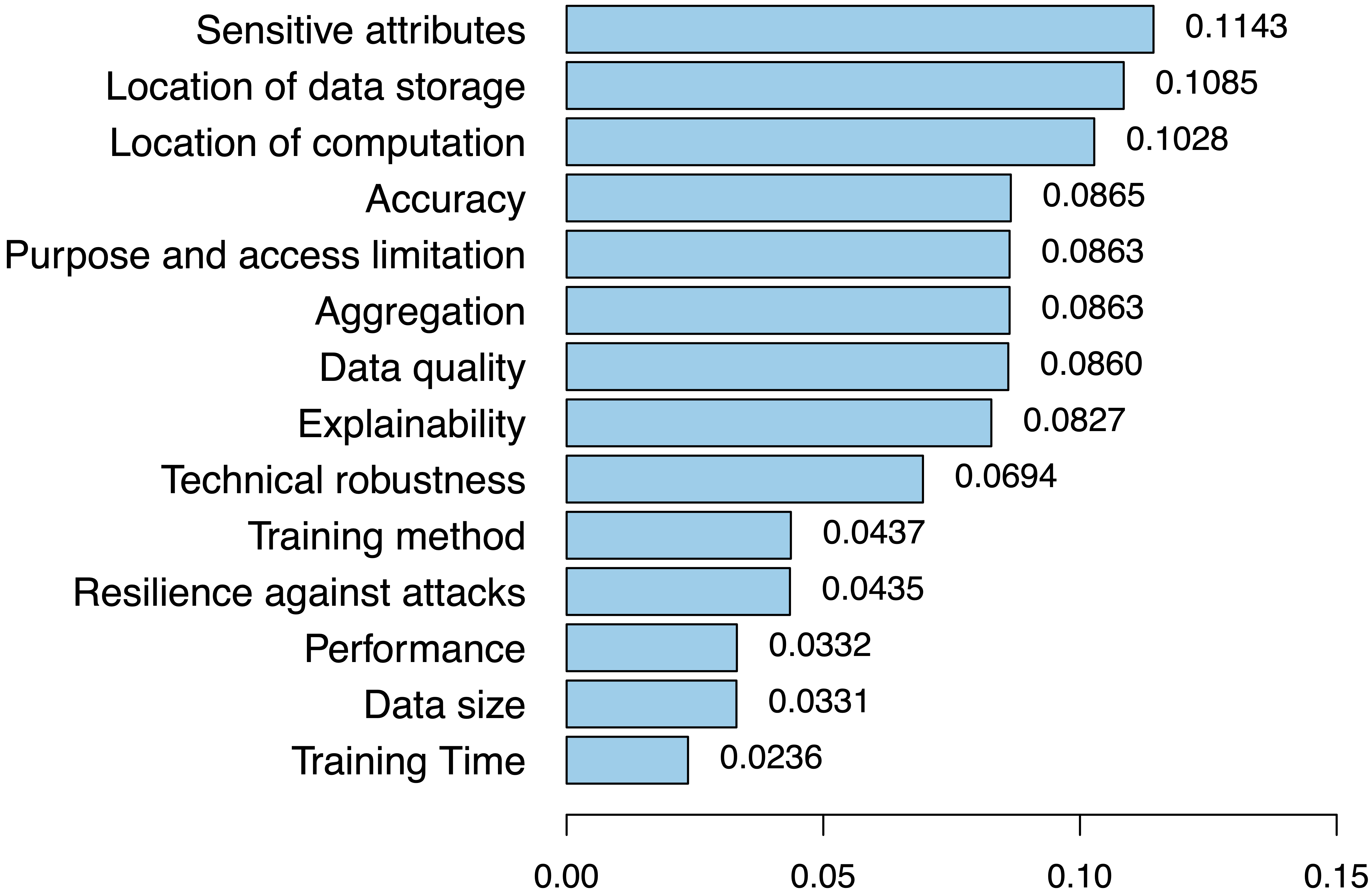}
          \caption{User $\tilde{c}_i$}\label{fig:PPML_scores_user}
    \end{subfigure}
    \hfill
     \begin{subfigure}[b]{0.50\textwidth}
\centering
\includegraphics[width=\textwidth]{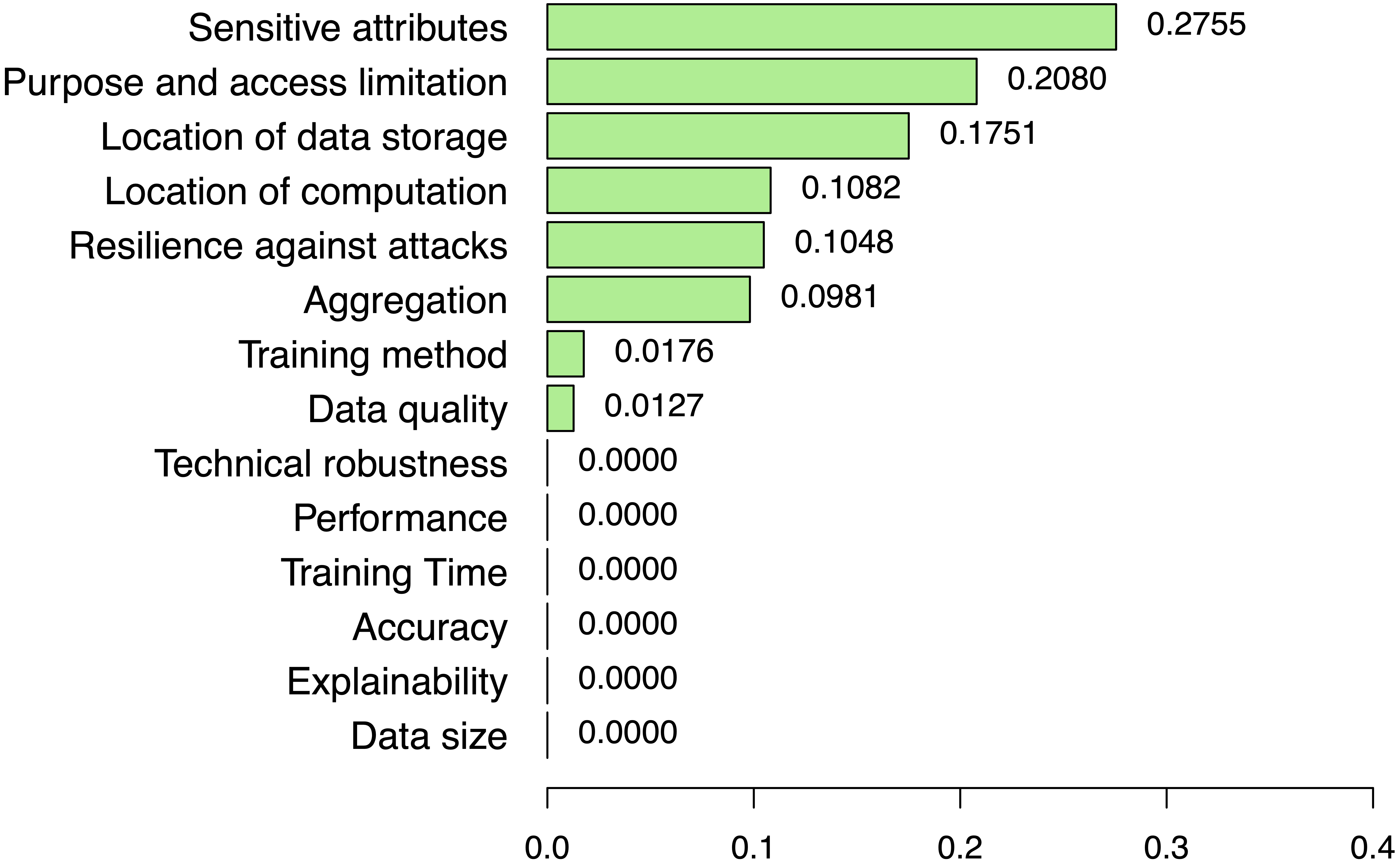}
\caption{Data entity $\tilde{c}_i$ \label{fig:PPML_scores_data_entity}}
\end{subfigure}
\caption{Importance scores $\tilde{c}_i$  (x-axis) of PPML Characteristics (y-axis) for user/data entity\label{fig:scores}}
\end{figure}

In this section, we derive the technology ranking based on the previous table by calculating a score that reflects the technology evaluation from table \ref{tab:PPML_metrics} and the PPML importance scores from figure\;\ref{fig:PPML_scores_user}.
\par
The first step in this calculation is to translate table \ref{tab:PPML_metrics} that represents the values of $X$ into numbers we can use for calculation. E.g., for the PPML Characteristics \textit{sensitive attributes} $X_2$ and $y_2$ are represented the following: 
\[
X_{2} =
\begin{bmatrix}
    0 & 0 & 0 & 0 & 0 \\
    0 & 1 & 1 & 0 & 0 \\
    0 & 0 & 0 & 1 & 1 \\
    0 & 0 & 0 & 0 & 0 \\
    1 & 1 & 1 & 0 & 0 \\
\end{bmatrix}
\quad
y_2 = \begin{bmatrix} 0.2 \\ 0.3 \\ 0.3 \\ 1 \\ 0.2 \end{bmatrix}
\]

The hard requirements have to be fulfilled and a technology will be removed if it does not. In a later implementation we plan to invoke a notification if a hard requirement is not fulfilled. Since we have only evaluated techniques that do fulfill the hard requirements we reduce I to 10. Regarding the trade-offs (see table \ref{tab:PPML_metrics}) we have set $T = 1$ because the trade-off was already considered when setting e.g. a lower expected accuracy for MDP.
We first derive the dot product of $X$ and $y$ (see table \ref{tab:xy}). Now we can compute the score $e$ using our formula that includes the user preference scores $\tilde{c_i}$, the technology characteristic evaluations $X_i$ and the weight vectors $y_i$: 

$$e := \sum_{i=0}^{I=10} \tilde{c_i} \cdot (X_i^T \cdot y_i)$$

\par
The final scores of $e$ are shown in table \ref{tab:e}. In addition we also show the intermediate results for each PPML Characteristic and Technology. These results already contain the category weights ($y$) and the users' normalised PPML Characteristic preferences ($\tilde{c_i}$).
\begin{table}[htb!]
    \centering
    \begin{footnotesize}
    \caption{Users' PPML Characteristic scores for each PPML technology: $(X_i \cdot y_i)$}
    \label{tab:xy}
    \begin{tabularx}{.7\textwidth}{l *{5}{>{\centering\arraybackslash}X}}
        \hline
        & \textbf{FL} & \textbf{FL+LDP} & \textbf{MDP} & \textbf{SMPC} & \textbf{HE} \\
        \hline
    Data Quality & 1.00 & 1.00 & 1.00 & 1.00 & 1.00 \\
    Aggregation & 0.25 & 0.25 & 1.00 & 1.00 & 1.00 \\
    Sensitive Attributes & 0.20 & 0.50 & 0.50 & 0.30 & 0.30 \\
    Explainability & 1.00 & 1.00 & 1.00 & 0.50 & 0.50 \\
    Location of Computation & 0.70 & 0.70 & 0.20 & 0.30 & 0.20 \\
    Accuracy F1 & 1.00 & 0.30 & 0.00 & 1.00 & 1.00 \\
    Training Time & 1.00 & 1.00 & 1.00 & 0.30 & 0.30 \\
    Performance & 0.60 & 0.60 & 0.60 & 0.60 & 0.30 \\
    Resilience Against Attacks & 0.00 & 0.50 & 0.50 & 1.00 & 1.00 \\
    Purpose and Access Limitation & 0.00 & 0.50 & 0.50 & 0.75 & 1.00 \\
    Technical Robustness & 1.00 & 1.00 & 1.00 & 0.00 & 0.00 \\ \hline
    SUM &6.75 & 7.35 &7.30&6.75&6.60 \\
        \hline
    \end{tabularx}
    \end{footnotesize}
\end{table}

\begin{table}[htb!]
    \centering
    \caption{Users' PPML Characteristic scores for each PPML technology: $\tilde{c_i} \cdot (X_i \cdot y_i)$}
    \label{tab:e}
    \begin{footnotesize}
    \begin{tabularx}{.9\textwidth}{l *{5}{>{\centering\arraybackslash}X}}
        \hline
        & \textbf{FL} & \textbf{FL+LDP} & \textbf{MDP} & \textbf{SMPC} & \textbf{HE} \\
        \hline
       

Data quality                   &  0.086036 & 0.086036 & 0.086036 & 0.086036 & 0.086036 \\
Aggregation                    &  0.021572 & 0.021572 & 0.086289 & 0.086289 & 0.086289 \\ 
Sensitive attributes           &  0.022865 & 0.057162 & 0.057162 & 0.034297 & 0.034297 \\
Explainability                 &  0.082736 & 0.082736 & 0.082736 & 0.041368 & 0.041368 \\
Location of computation        &  0.071946 & 0.071946 & 0.020556 & 0.030834 & 0.020556 \\
Accuracy F1                    &  0.086531 & 0.025959 & 0.000000 & 0.086531 & 0.086531 \\
Training time                  &  0.023646 & 0.023646 & 0.023646 & 0.007094 & 0.007094 \\
Performance                    &  0.019897 & 0.019897 & 0.019897 & 0.019897 & 0.009949 \\
Resilience against attacks     &  0.000000 & 0.021752 & 0.021752 & 0.043504 & 0.043504 \\
Purpose and access limitation  &  0.000000 & 0.043144 & 0.043144 & 0.064716 & 0.086289 \\
Technical robustness           &  0.069414 & 0.069414 & 0.069414 & 0.000000 & 0.000000 \\
\hline
    
    $e$ & 0.484643 & 0.523264 & 0.510632 & 0.500566 & 0.501913 \\
        \hline
    \end{tabularx}
    \end{footnotesize}
\end{table}

Comparing the final results we see that FL+LDP has achieved the highest score. Comparing the PPML Characteristic scores, we notice that no technology is dominating another. 
Comparing FL and FL+LDP the addition of LDP increases the score for \textit{Resilience against attacks} and \textit{Purpose and access limitations} (see table \ref{tab:xy}). The cost of this is reduced accuracy, evaluated by a lower F1-Score. This is not surprising and comes from the fact that DP in general is likely to decrease the accuracy in favour of privacy. This is also notable when comparing FL with MDP. The biggest difference in these architectures is that LDP is used locally on the gradients of the FL architecture while MDP is applied directly on the data at a central entity. The highest scores in privacy and security characteristics were achieved by SMPC and HE while HE has the best scores. 
Comparing the results in table \ref{tab:xy} and table \ref{tab:e} that takes the user weights into account the relative importance of SMPC has slightly decreased compared to HE. The reason for this is that \textit{performance} where SMPC is superior was assessed less important (see figure \ref{fig:PPML_scores_user}) with 0.0332 compared to the other PPML characteristics. The score of FL+LDP was increased compared to the second best solution because location of \textit{computation} has a high importance for the users with 0.1085.
For completeness and the sake of showcasing we have also used the potential user preferences to compute the data entities' results that are presented in table \ref{tab:entity}. In this table, we have removed all rows that do not have an impact on PPML Characteristics based on table \ref{tab:mapping} and would thus just exhibit the value 0 (compare figure \ref{fig:PPML_scores_data_entity}). We find that the preference for FL and FL+LDP are the lowest for the data entity compared to the users. For a more comprehensive statement it might be useful to collect the preferences of data entities directly when collecting their data. This is more realistic than using data of potential users and can help to avoid bias in data. 
\begin{table}[t!]
\centering
    \caption{Data entities' PPML Characteristic scores for each PPML technology: $\tilde{c_i} \cdot (X_i \cdot y_i)$.}\label{tab:entity}
    \begin{footnotesize}
    \begin{tabularx}{.9\textwidth}{l *{5}{>{\centering\arraybackslash}X}}
        \hline
        & \textbf{FL} & \textbf{FL+LDP} & \textbf{MDP} & \textbf{SMPC} & \textbf{HE} \\
        \hline

Data Quality                  & 0.012707 & 0.012707 & 0.012707 & 0.012707 & 0.012707 \\
Aggregation                   & 0.024529 & 0.024529 & 0.098116 & 0.098116 & 0.098116 \\
Sensitive Attributes          & 0.055104 & 0.137760 & 0.137760 & 0.082656 & 0.082656 \\
Location of Computation       & 0.075706 & 0.075706 & 0.021630 & 0.032445 & 0.021630 \\
Resilience Against Attacks    & 0.000000 & 0.052423 & 0.052423 & 0.104845 & 0.104845 \\
Purpose and Access Limitation & 0.000000 & 0.103977 & 0.103977 & 0.155966 & 0.207955 \\

    \hline
    $e$ & 0.168046 & 0.407102  & 0.426613 & 0.486735 & 0.527909 \\
    
    \hline
\end{tabularx}\end{footnotesize}
\end{table} 
\par

How to weight between data entities and users and which technology is finally chosen is not trivial since further features can impact the decision, e.g. maintenance costs. Still, our framework provides detailed insights in most important PPML Characteristics that satisfy the users' needs.

\section{Discussion}\label{sec:discussion}
In this section we discuss our impact, lessons learned and future work.

\subsection{Impact}
With the rise of the AI Act, the need for PPML has been confirmed by the European Commission. Incumbents that are in charge to implement privacy for their AI application as well as new entering companies can utilise our framework to understand the needs of their customers and thus adjust the privacy strategies. We aim to strengthen the principles of privacy by design by providing clarification on which PPML technology will bring the highest user satisfaction. This aims to ensure full functionality while providing a high privacy level. Regarding research we identified a gap in structured comparison of PPML technologies that take user preferences into account. We hope to encourage other researchers to close this gap.

\subsection{Empowering of user rights}
With this work we empower the user of an AI service with the possibility to choose a technology that meets best their preferences. Although the users are by our approach technically empowered to choose the PPML with the best security and privacy, this does not necessarily mean that they do so. Nevertheless, optimizing efficiency of an application by trading e.g. performance against privacy is an intended possible outcome of the framework. We follow the approach that security and privacy in an application is only useful if the application still provides full functionality \citep{cavoukian2009privacy}.

\subsection{Lessons learned}

During the evaluation of the categories we noticed that dependencies between categories exist. E.g., for DP often a trade-off between accuracy and privacy is reported. While we take this into account with a reduced weight for these dependencies, we will further investigate how to best implement trade-offs in an advanced version of the framework.
\par
To implement user preferences for a PPML technology decision requires many steps and specific technological expertise. How to improve the evaluation time is a topic we will address in future.
\par
Since we have experimented with a lot of different UAC rankings we noticed that small changes in the UAC preference scores can have a huge impact on  the final result. Thus, robustness of the model should be investigates with further studies in future. 
\par
When analysing the different steps to derive the framework results we had several feedback loops to ensure that there is notable impact of UAC preference scores on the final results. To get a good balance between technology and UAC preference score influence is not an easy task. In our future work, we want to dive deeper into the explainability and tracing of influences.

\subsection{Limitations}
While the framework is a first step towards the possibility to consider user preferences for the selection of a PPML technology, it requires parameters specific to each considered scenario. For example, some of the thresholds and weights for PPML Characteristics (cf. Table~\ref{tab:PPML_metrics}) might be different for another scenario. Determining these parameters can take a considerable amount of effort and requires expertise in several disciplines.

\par
Regarding the application of the framework in the PSI detection use case, we had to rely on the statements from existing literature, e.g., regarding privacy guarantees or run-times of the different technologies. It is important to be aware of the setup differences in the literature and resulting reported accuracy and performance disparities. 
\par
The values in table \ref{table:y_justification} are an example and might be influenced by personal preferences since it is can be hard to come up with reasonable values and scales, especially if literature is scarce. 
\par
Another limitation is the number of participants in the AHP that we surveyed to calculate the UAC preference scores. However, since we were only interested in working through the use case as a demonstration of our framework, the impact is negligible.

\subsection{Future Work}
Since we have just used a small group of people participating in the AHP we will conduct further research on how to best evaluate our UAC. We aim to evaluate a different use case with a larger participant-number. 
In future, we also want to further improve the connections in the mapping itself based on the experts' feedback. Thus we plan additional interviews with PPML experts to evaluate the whole framework.
While we have now implemented the framework for a B2C use case, we are curious to investigate how the framework can be extended to B2B use cases. 
Therefore, our future main objective is an expert evaluated, advanced framework that is less time consuming.

\section{Conclusion}\label{sec:conclusion}
In this paper, we have provided a process to systematically include user preferences for privacy preserving machine learning techniques (PPMLs) without asking for technical details. To do so we have suggested a framework that takes user preferences and developer knowledge as input to calculate a weighted score for each PPML. After providing the theoretical approach, we have applied the framework to a simplified PSI detection use case. While our previous work has focused on the mapping of User Acceptance Criteria (UAC) and PPML characteristics, this line of research focused on the collection of developer input and the sequential operation steps to calculate the weighted scores for each PPML technique.


\section*{Acknowledgements}
This research received funding by the European Union (EU) - NextGenerationEU, and the German Federal Ministry of Education and Research under the BMBF project FIIPS-at-Home. The views expressed are those of the author(s) and do not necessarily reflect those of the EU or the EC.

\bibliographystyle{plainnat}
\bibliography{bib}

\end{document}